\documentclass[letterpaper, 10 pt, conference]{ieeeconf}  

\IEEEoverridecommandlockouts                              

\usepackage[T1]{fontenc}
\usepackage[latin9]{luainputenc}

\usepackage{amsmath}
\usepackage{amssymb}
\usepackage{graphicx}
\makeatletter
\usepackage{lmodern}

\usepackage[T1]{fontenc}
\usepackage{courier}
\usepackage{color}
\usepackage{upquote}
\usepackage{xcolor}
\usepackage{listings}
\usepackage{tikzscale}
\usepackage{pgfplots}
\pgfplotsset{compat=1.16}
\usepackage{tikz}

\usepackage{bm}
\usepackage{mathtools}
\usepackage{graphicx, epstopdf}
\usepackage[framed, numbered]{matlab-prettifier}
\usepackage{dsfont}
\usepackage{paralist}
\usepackage[nolist]{acronym}
\usepackage{algorithm,algorithmic}

\definecolor{hellgelb}{rgb}{1,1,0.85} 
\definecolor{colKeys}{rgb}{0,0,1} 
\definecolor{colIdentifier}{rgb}{0,0,0} 
\definecolor{colComments}{rgb}{0,0.5,0} 
\definecolor{colString}{rgb}{0.81,0.12,0.95}
\newcommand{\CVaR}{\mathrm{CVaR}}

\newcommand{\Tr}{\mathrm{T}} 

\DeclareMathOperator*{\argmin}{arg\,min}

\usepackage{xcolor}

\newcommand{\blind}[1]{{\color{black} #1 }} 

\newcommand{\algorithmicbreak}{\textbf{break}}
\newcommand{\BREAK}{\STATE \algorithmicbreak}

\usepackage{microtype}

\renewcommand{\baselinestretch}{0.97}
\renewcommand{\baselinestretch}{0.982}

\usepackage{setspace}

\usepackage[subtle]{savetrees}

\usepackage{biblatex}
\usepackage{hyperref}

\hypersetup{
    colorlinks=false,
    linkcolor=blue,
    filecolor=magenta,      
    urlcolor=cyan,
}

\title{\LARGE \bf
STEP: Stochastic Traversability Evaluation and Planning for Risk-Aware Off-road Navigation
}

\author{
David D. Fan$^{*1}$, Kyohei Otsu$^{*1}$, Yuki Kubo$^{2}$, Anushri Dixit$^{3}$,\\
Joel Burdick$^{3}$, and Ali-Akbar Agha-Mohammadi$^{1}$
\thanks{$^*$These authors contributed equally to this work.}
\thanks{$^{1}$These authors are with the Jet Propulsion Laboratory, California Institute of Technology, Pasadena, CA
        \tt\{david.d.fan,kyohei.otsu,aliagha\}@jpl.nasa.gov}%
\thanks{$^{2}$This author is with the University of Tokyo, Tokyo, Japan
        \tt kubo.yuki@ac.jaxa.jp}%
\thanks{$^{3}$These authors are with California Institute of Technology, Pasadena, CA
        \tt\{adixit,jwb\}@robotics.caltech.edu}%
\thanks{\textcopyright 2021, California Institute of Technology. All Rights Reserved}%
}

\begin{document}

\maketitle
\begin{abstract}
Although ground robotic autonomy has gained widespread usage in structured and controlled environments, autonomy in unknown and off-road terrain remains a difficult problem.  Extreme, off-road, and unstructured environments such as undeveloped wilderness, caves, and rubble pose unique and challenging problems for autonomous navigation.  To tackle these problems we propose an approach for assessing traversability and planning a safe, feasible, and fast trajectory in real-time.  Our approach, which we name STEP (Stochastic Traversability Evaluation and Planning), relies on: 1) rapid uncertainty-aware mapping and traversability evaluation, 2) tail risk assessment using the Conditional Value-at-Risk (CVaR), and 3) efficient risk and constraint-aware kinodynamic motion planning using sequential quadratic programming-based (SQP) model predictive control (MPC).  We analyze our method in simulation and validate its efficacy on wheeled and legged robotic platforms exploring extreme terrains including an abandoned subway and an underground lava tube. (See video: \url{https://youtu.be/N97cv4eH5c8})
\end{abstract}


\section{Introduction}

\begin{figure}[t!]
    \centering
    \includegraphics[width=\linewidth]{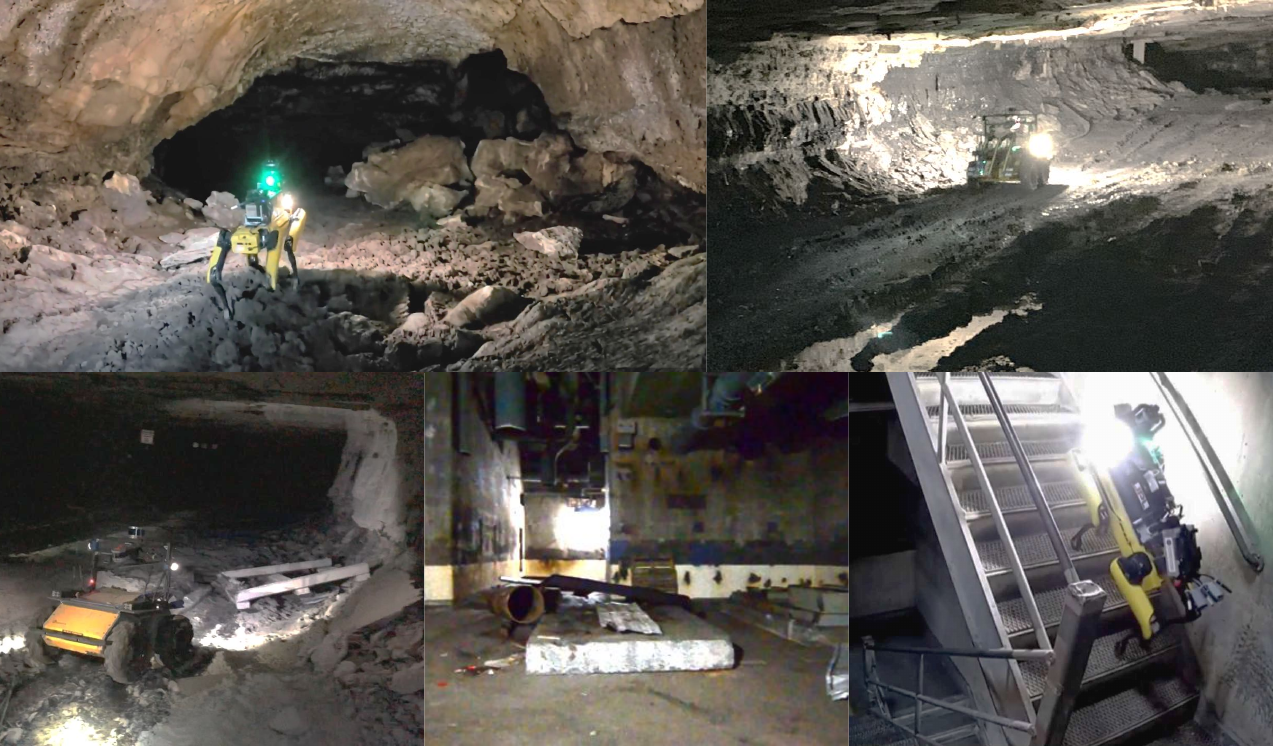}
    \caption{Top left: Boston Dynamics Spot quadruped robot exploring Valentine Cave at Lava Beds National Monument, CA.  Top right, bottom left:  Clearpath Husky robot exploring Arch Mine in Beckley, WV.  Bottom middle, right:  Spot exploring abandoned Satsop power plant in Elma, WA.}
    \label{fig:HuskyArch}
\end{figure}

Consider the problem of a ground robot tasked to autonomously traverse an unknown environment.  In real-world scenarios, environments which are of interest to robotic operations are highly risky, containing difficult geometries (e.g. rubble, slopes) and non-forgiving hazards (e.g. large drops, sharp rocks) (See Figure \ref{fig:HuskyArch}) \cite{kalita2018path}.  Determining where the robot may safely travel is a non-trivial problem, compounded by several issues:  1) Localization error affects how sensor measurements are accumulated to generate dense maps of the environment. 2) Sensor noise, sparsity, and occlusion induces biases and uncertainty in analysis of traversability. 3)  Environments often pose a mix of various sources of traversability risk, including slopes, rough terrain, low traction, narrow passages, etc.  4)  These various risks create highly non-convex constraints on the motion of the robot, which are compounded by the kinodynamic constraints of the robot itself.

\begin{figure} 
\centering
    \resizebox{0.45\textwidth}{!}{%
\begin{tikzpicture}
\tikzstyle{every node}=[font=\large]
\pgfmathdeclarefunction{gauss}{2}{%
  \pgfmathparse{1000/(#2*sqrt(2*pi))*((x-.5-8)^2+.5)*exp(-((x-#1-6)^2)/(2*#2^2))}%
}

\pgfmathdeclarefunction{gauss2}{3}{%
\pgfmathparse{1000/(#2*sqrt(2*pi))*((#1-.5-8)^2+.5)*exp(-((#1-#1-6)^2)/(2*#2^2))}
}

\begin{axis}[
  no markers, domain=0:16, range=-2:8, samples=200,
  axis lines*=center, xlabel=$\zeta$, ylabel=$p(\zeta)$,
  every axis y label/.style={at=(current axis.above origin),anchor=south},
  every axis x label/.style={at=(current axis.right of origin),anchor=west},
  height=5cm, width=17cm,
  xtick={0,5,7,10}, ytick=\empty,
  xticklabels={$0$, , , ,},
  enlargelimits=true, clip=false, axis on top,
  grid = major
  ]
  \addplot [fill=cyan!20, draw=none, domain=7:15] {gauss(1.5,2)} \closedcycle;
  \addplot [very thick,cyan!50!black] {gauss(1.5,2)};
 
 \pgfmathsetmacro\valueA{gauss2(5,1.5,2)}
 \draw [gray] (axis cs:5,0) -- (axis cs:5,\valueA);
  \pgfmathsetmacro\valueB{gauss2(10,1.5,2)}
  \draw [gray] (axis cs:4.5,0) -- (axis cs:4.5,\valueB);
    \draw [gray] (axis cs:10,0) -- (axis cs:10,\valueB);
 
 
 \draw [gray] (axis cs:1,0)--(axis cs:5,0);
 \node[below] at (axis cs:7.0, -0.1)  {$\mathrm{VaR}_{\alpha}(\zeta)$}; 
\node[below] at (axis cs:5, -0.1)  {$\mathbb{E}(\zeta)$}; 
\node[below] at (axis cs:10, -0.1)  {$\mathrm{CVaR}_{\alpha}(\zeta)$};
\draw [yshift=2cm, latex-latex](axis cs:7,0) -- node [fill=white] {Probability~$1-\alpha$} (axis cs:16,0);
\end{axis}
\end{tikzpicture}
}
\caption{Comparison of the mean, VaR, and CVaR for a given risk level $\alpha \in (0,1]$. The axes denote the values of the stochastic variable $\zeta$, which in our work represents traversability cost. The shaded area denotes the $(1-\alpha)\%$ of the area under $p(\zeta)$.  $\CVaR_{\alpha}(\zeta)$ is the expected value of $\zeta$ under the shaded area.}
\label{fig:varvscvar}
\end{figure}
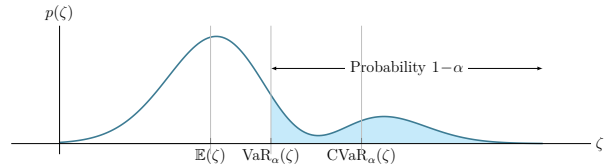

To address these issues we adopt an approach in which we directly quantify the traversal cost along with the uncertainties associated with that cost.  We refer to this cost as \textit{traversability}, e.g. a region of the environment in which the robot will suffer or become damaged has a high traversability cost.  \blind{Building upon our previous work on traversability in extreme terrains \cite{thakker2021autonomous},} 
we formulate the problem as a risk-aware, online nonlinear Model Predictive Control (MPC) problem, in which the uncertainty of traversability is taken into account when planning a trajectory.  Our goal is to minimize the traversability cost, but directly minimizing the mean cost leads to an unfavorable result because tail events with low probability of occurrence (but high consequence) are ignored (Figure \ref{fig:varvscvar}).  Instead, in order to quantify the impact of uncertainty and risk on the motion of the robot, we employ a formulation in which we find a trajectory which minimizes the Conditional Value-at-Risk (CVaR) \cite{rockafellar2000optimization}.  Because CVaR captures the expected cost of the tail risk past a given probability threshold, we can dynamically adjust the level and severity of uncertainty and risk we are willing to accept (which depends on mission-level specifications, user preference, etc.).  While online chance-constrained nonlinear MPC problems often suffer from a lack of feasibility, our approach allows us to relax the severity of CVaR constraints by adding a penalizing loss function.

We quantify risk via a traversability analysis pipeline (for system architecture, see Figure \ref{fig:system}).  At a high level, this pipeline creates an uncertainty-aware 2.5D traversability map of the environment by aggregating uncertain sensor measurements.  Next, the map is used to generate both environment and robot induced costs and constraints.  These constraints are convexified and used to build an online receding horizon MPC problem, which is solved in real-time.  As we will demonstrate, we push the state-of-the-art in making this process highly efficient, allowing for re-planning at rates which allow for dynamic responses to changes and updates in the environment, as well as high travel speeds.

In this work, we propose STEP (Stochastic Traversability Evaluation and Planning), that pushes the boundaries of the state-of-the-practice to enable safe, risk-aware, and high-speed ground traversal of unknown environments.  Specifically, our contributions include:
\begin{enumerate}
    \item Uncertainty-aware 2.5D traversability evaluation which accounts for localization error, sensor noise, and occlusion, and combines multiple sources of traversability risk.
    \item An approach for combining these traversability risks into a unified risk-aware CVaR planning framework.
    \item A highly efficient MPC architecture for robustly solving non-convex risk-constrained optimal control problems.
    \item Real-world demonstration of real-time CVaR planning on wheeled and legged robotic platforms in unknown and risky environments.
\end{enumerate}


\section{Related Work}
Our work is related to other classical approaches to traversability.  Most traversability analyses are dependent on sensor type and measured through geometry-based, appearance-based, or proprioceptive methods \cite{papdakis2013survey}. Geometry-based methods often rely on building a 2.5D terrain map which is used to extract features such as maximum, minimum, and variance of the height and slope of the terrain \cite{gestalt}. Planning algorithms for such methods take into account the stability of the robot on the terrain \cite{hait2002algorithms}. In \cite{otsu2020fast,ghosh2018pace}, the authors estimate the probability distributions of states based on the kinematic model of the vehicle and the terrain height uncertainty.  Furthermore, a method for incorporating sensor and state uncertainty to obtain a probabilistic terrain estimate in the form of a grid-based elevation map was considered in \cite{fankhauser2018probabilisticterrain}.  Our work builds upon these ideas by performing traversability analyses using classical geometric methods, while incorporating the uncertainty of these methods for risk-aware planning \cite{agha2017CRM, thakker2021autonomous}.

Risk can be incorporated into motion planning using a variety of different methods, including chance constraints \cite{ono2015chance,wang2020non}, exponential utility functions \cite{koenig1994risk}, distributional robustness \cite{xu2010distributionally}, and quantile regression \cite{fan2020deep,dabney2018distributional}. Risk measures, often used in finance and operations research, provide a mapping from a random variable (usually the cost) to a real number.  These risk metrics should satisfy certain axioms in order to be well-defined as well as to enable practical use in robotic applications \cite{majumdar2020should}.  Conditional value-at-risk (CVaR) is one such risk measure that has this desirable set of properties, and is a part of a class of risk metrics known as \textit{coherent risk measures} \cite{artzner1999coherent}
Coherent risk measures have been used in a variety of decision making problems, especially Markov decision processes (MDPs) \cite{chow2015risk}. In recent years, Ahmadi et al. synthesized risk averse optimal policies for partially observable MDPs, constrained MDPs, and for shortest path problems in MDPs \cite{ahmadi2020uncertainty, ahmadi2020constrained, ahmadi2021riskaverse}.  
Coherent risk measures have been used in a MPC framework when the system model is uncertain \cite{singh2018framework} and when the uncertainty is a result of measurement noise or moving obstacles \cite{dixit2020risksensitive}.  
In \cite{hakobyan2019risk, dixit2020risksensitive}, the authors incorporated risk constraints in the form of distance to the randomly moving obstacles but did not include model uncertainty.  Our work extends CVaR risk to a risk-based planning framework which utilizes different sources of traversability risk (such as collision risk, step risk, slippage risk, etc.) Morever, this paper introduces the first field-hardened and theoretically grounded approach to traversability assessment and risk-constrained planning using CVaR metrics. Using CVaR to assess traversability risks allows us to dynamically tune the entire system's behavior - from aggressive to highly conservative - by changing a single value, the risk probability level.

Model Predictive Control has a long history in controls as a means to robustly control more complex systems, including time-varying, nonlinear, or MIMO systems \cite{camacho2013model}.  While simple linear PID controllers are sufficient for simpler systems, MPC is well-suited to more complex tasks while being computationally feasible.  In this work, MPC is needed to handle a) complex interactions (risk constraints) between the robot and the environment, including non-linear constraints on robot orientation and slope, b) non-linear dynamics which include non-holonomic constraints, and c) non-convex, time-varying CVaR-based constraints and cost functions.  In particular, we take an MPC approach known as Sequential Quadratic Programming (SQP), which iteratively solves locally quadratic sub-problems to converge to a globally (more) optimal solution \cite{boggs1995sequential}.  Particularly in the robotics domain, this approach is well-suited due to its reduced computational costs and flexibility for handling a wide variety of costs and constraints \cite{schulman2014motion, lew2020chance}.  A common criticism of SQP-based MPC (and nonlinear MPC methods in general) is that they can suffer from being susceptible to local minima.  We address this problem by incorporating a trajectory library (which can be predefined and/or randomly generated, e.g. as in \cite{kalakrishnan2011stomp}) to use in a preliminary trajectory selection process.  We use this as a means to find more globally optimal initial guesses for the SQP problem to refine locally.  Another common difficulty with risk-constrained nonlinear MPC problems is ensuring recursive feasibility \cite{lofberg2012oops}.  We bystep this problem by dynamically relaxing the severity of the risk constraints while penalizing CVaR in the cost function.

\section{Risk-Aware Traversability and Planning}
\subsection{Problem Statement}
We first give a formal definition of the problem of risk-aware traversability and motion planning.
%
Let $x_k$, $u_k$, $z_k$ denote the state, action, and observation at the $k$-th time step. A path $x_{0:N}=\{x_0, x_1, \cdots, x_N\}$ is composed of a sequence of poses. A policy is a mapping from state to control $u = \pi(x)$. A map is represented as $m = (m^{(1)}, m^{(2)}, \cdots)$ where $m^{i}$ is the $i$-th element of the map (e.g., a cell in a grid map).
%
The robot's dynamics model captures the physical properties of the vehicle's motion, such as inertia, mass, dimension, shape, and kinematic and control constraints:
\begin{align}
    x_{k+1} &= f(x_{k}, u_{k}) \\
    g(u_{k}) &\succ 0
\end{align}
where $g(u_k)$ is a vector-valued function which encodes control constraints/limits.

Following \cite{papdakis2013survey}, we define \textit{traversability} as the capability for a ground vehicle to reside over a terrain region under an admissible state.  We represent traversability as a cost, i.e. a continuous value computed using a terrain model, the robotic vehicle model, and kinematic constraints, which represents the degree to which we wish the robot to avoid a given state:
\begin{align}
    r = \mathcal{R}(m, x, u)
\end{align}
where $r\in\mathbb{R}$, and $\mathcal{R}(\cdot)$ is a traversability assessment model.  This model captures various unfavorable events such as collision, getting stuck, tipping over, high slippage, to name a few. Each mobility platform has its own assessment model to reflect its mobility capability.

Associated with the true traversability value is a distribution over possible values based on the current understanding about the environment and robot actions. In most real-world applications where perception capabilities are limited, the true value can be highly uncertain.  To handle this uncertainty, consider a map belief, i.e., a probability distribution $p(m | x_{0:k}, z_{0:k})$, over a possible set $\mathcal{M}$.  Then, the traversability estimate is also represented as a random variable $R: (\mathcal{M} \times \mathcal{X} \times \mathcal{U}) \longrightarrow \mathbb{R}$.  We call this probabilistic mapping from map belief, state, and controls to possible traversability cost values a \textit{risk assessment model}.

A risk metric $\rho(R):R\rightarrow\mathbb{R}$ is a mapping from a random variable to a real number which quantifies some notion of risk.  In order to assess the risk of traversing along a path $x_{0:N}$ with a policy $\pi$, we wish to define the cumulative risk metric associated with the path, $J(x_0,\pi)$.  To do this, we need to evaluate a sequence of random variables $R_{0:N}$.  To quantify the stochastic outcome as a real number, we use the dynamic, time-consistent risk metric given by compounding the one-step risk metrics \cite{ruszczynski2014riskaverseDP}:
\begin{align}
  J(x_0, \pi; m)
  &= R_0 + \rho_0\big( R_1 + \rho_{1}\big(R_2 + \dotsc + \rho_{N-1}\big(R_{N})\big)\big)
  \label{eq:risk_metric}
\end{align}
where $\rho_k(\cdot)$ is a one-step coherent risk metric at time $k$. This one-step risk gives us the cost incurred at time-step $k+1$ from the perspective of time-step $k$. Any distortion risk metric compounded as given in \eqref{eq:risk_metric} is time-consistent (see \cite{majumdar2020should} for more information on distortion risk metrics and time-consistency). We use the Conditional Value-at-Risk (CVaR) as the one-step risk metric:
\begin{align}
    \rho(R) =
    \mathrm{CVaR}_{\alpha}(R)=
    &\inf_{z \in \mathbb{R}} \mathbb{E}\Bigg[z + \frac{(R-z)_{+}}{1-\alpha}\Bigg]
\end{align}
where $(\cdot)_{+}=\max(\cdot, 0)$, and $\alpha \in (0, 1]$ denotes the \textit{risk probability level}.

We formulate the objective of the problem as follows: Given the initial robot configuration $x_{S}$ and the goal configuration $x_{G}$, find an optimal control policy $\pi^{*}$ that moves the robot from $x_{S}$ to $x_{G}$ while 1) minimizing time to traverse, 2) minimizing the cumulative risk metric along the path, and 3) satisfying all kinematic and dynamic constraints.

\subsection{Hierarchical Risk-Aware Planning}



We propose a hierarchical approach to address the aforementioned risk-aware motion planning problem by splitting the motion planning problem into geometric and kinodynamic domains.  We consider the geometric domain over long horizons, while we solve the kinodynamic problem over a shorter horizon. 
This is convenient for several reasons:  1) Solving the full constrained CVaR minimization problem over long timescales/horizons becomes intractable in real-time.  2) Geometric constraints play a much larger role over long horizons, while kinodynamic constraints play a much larger role over short horizons (to ensure dynamic feasibility at each timestep).  3) A good estimate (upper bound) of risk can be obtained by considering position information only.  This is done by constructing a position-based traversability model $\mathcal{R}_{\mathrm{pos}}$ by marginalizing out non-position related variables from the risk assessment model, i.e. if the state $x=[p_x,p_y,x_{\mathrm{other}}]^{\intercal}$ consists of position and non-position variables (e.g. orientation, velocity), then
\begin{align}
        \mathcal{R}_{\mathrm{pos}}(m,p_x,p_y) \geq \mathcal{R}(m,x,u) \quad \forall x_{\mathrm{other}}, u
\end{align}


\textit{Geometric Planning:} The objective of geometric planning is to search for an \textit{optimistic} risk-minimizing path, i.e. a path that minimizes an upper bound approximation of the true CVaR value.  For efficiency, we limit the search space only to the geometric domain.  We are searching for a sequence of poses $x_{0:N}$ which ends at $x_G$ and minimizes the position-only risk metric in \eqref{eq:risk_metric}, which we define as $J_{\mathrm{pos}}(x_{0:N})$.  The optimization problem can be written as:
\begin{align}
    x_{0:N}^{*} = \argmin_{x_{0:N}} \bigg[ J_{\mathrm{pos}}(x_{0:N}) &+ \lambda \sum_{k=0}^{N-1} \|x_k - x_{k+1}\|^2 \bigg]\label{eq:optimization_geometric_cost}\\
    s.t. \quad \phi(m, x_k) &\succ 0
    \label{eq:optimization_geometric_constraint}
\end{align}
where the constraints $\phi(\cdot)$ encode position-dependent traversability constraints (e.g. constraining the vehicle to prohibit lethal levels of risk) and $\lambda\in\mathbb{R}$ weighs the tradeoff between risk and path length.




\begin{figure*}[t!]
    \centering
    \includegraphics[width=\textwidth]{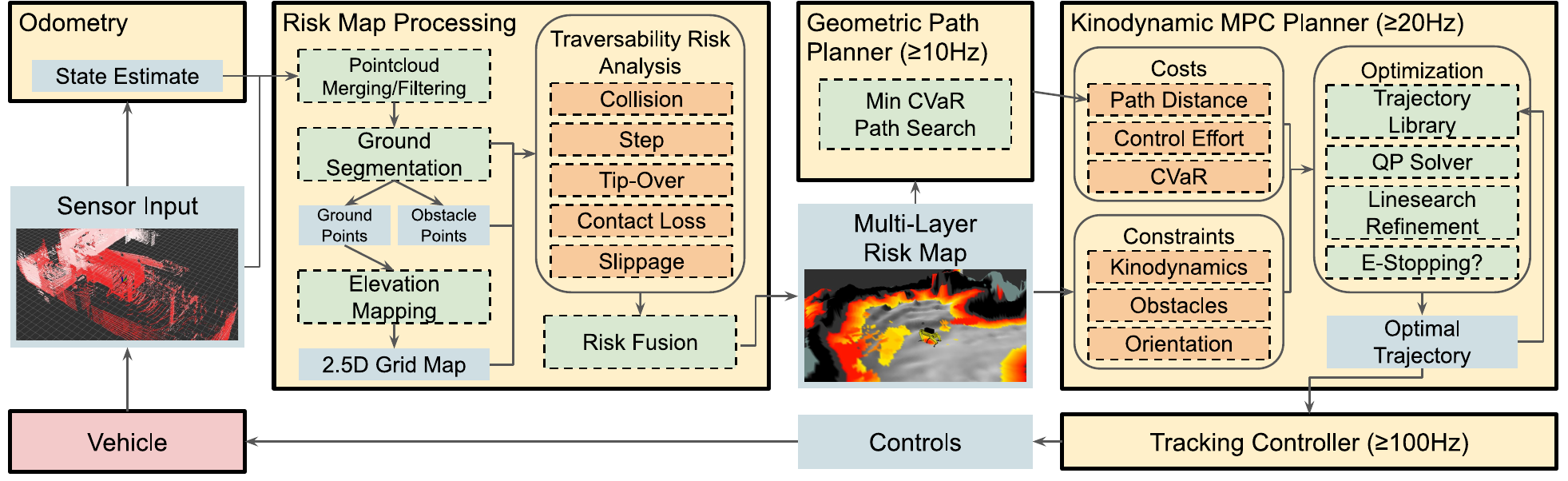}
    \caption{Overview of system architecture for STEP.  From left to right:  Odometry aggregates sensor inputs and relative poses.  Next, Risk Map Processing merges these pointclouds and creates a multi-layer risk map.  The map is used by the Geometric Path Planner and the Kinodynamic MPC Planner.  An optimal trajectory is found and sent to the Tracking Controller, which produces control inputs to the robot.}
    \label{fig:system}
\end{figure*}

\textit{Kinodynamic Planning:} We then solve a kinodynamic planning problem to track the optimal geometric path, minimize the risk metric, and respect kinematic and dynamics constraints. The goal is to find a control policy $\pi^{*}$ within a local planning horizon $T\leq N$ which tracks the path $X^*_{0:N}$.  The optimal policy can be obtained by solving the following optimization problem:

\begin{align}
 \pi^{*} = \argmin_{\pi \in \Pi} \bigg[
    J(x_0, \pi) + &\lambda \sum_{k=0}^T \|x_{k} - x^*_{k}\|^2
    \bigg]
    \label{eq:mpc_cost}\\
 s.t. ~ \forall k\in[0,\cdots,T]: \qquad x_{k+1} &= f(x_{k}, u_{k}) \label{eq:mpc_dynamics}\\ 
    g(u_k) &\succ 0 \label{eq:mpc_control_constraint}\\ 
    h(m, x_k) &\succ 0 
    \label{eq:mpc_state_constraints}
\end{align}
where the constraints $g(u)$ and $h(m, x_k)$ are vector-valued functions which encode controller limits and state constraints, respectively.

\section{STEP for Unstructured Terrain}

Having outlined our approach for solving the constrained CVaR minimization problem, in this section we discuss how we compute traversability risk and efficiently solve the risk-aware trajectory optimization problem.  At a high level, our approach takes the following steps (see Figure \ref{fig:system}):  1)  Assuming some source of localization with uncertainty, aggregate sensor measurements to create an uncertainty-aware map.  2)  Perform ground segmentation to isolate the parts of the map the robot can potentially traverse.  3)  Compute risk and risk uncertainty using geometric properties of the pointcloud (optionally, include other sources of risk, e.g. semantic or other sensors).  4) Aggregate these risks to compute a 2.5D CVaR risk map.  5)  Solve for an optimistic CVaR minimizing path over long ranges with a geometric path planner.  7) Solve for a kinodynamically feasible trajectory which minimizes CVaR while staying close to the geometric path and satisfying all constraints.

\subsection{Modeling Assumptions}

Among many representation options for rough terrain, we use a 2.5D grid map in this paper for its efficiency in processing and data storage \cite{Fankhauser2016GridMapLibrary}. The map is represented as a collection of terrain properties (e.g., height, risk) over a uniform grid.

For different vehicles we use different robot dynamics models.  For example, for a system which produces longitudinal/lateral velocity and steering (e.g. legged platforms), the state and controls can be specified as:
\begin{align}
    x &= [p_x, p_y, p_\theta, v_x, v_y, v_{\theta}]^{\intercal} \\
    u &= [a_x, a_y, a_{\theta}]^{\intercal}
\end{align}
While the dynamics $x_{k+1} = f(x_k,u_k)$ can be written as
$x_{k+1} = x_k + \Delta t \Delta x_k$, where $\Delta x_k = [v_x \cos(p_\theta) - v_y \sin(p_\theta), ~v_x \sin(p_\theta) + v_y \cos(p_\theta), ~\kappa v_x + (1-\kappa) v_{\theta},~ a_x,~ a_y, ~a_{\theta}]^\intercal$. 
We let $\kappa\in[0,1]$ be a constant which adjusts the amount of turning-in-place the vehicle is permitted.  In differential drive or ackermann steered vehicles we can remove the lateral velocity component of these dynamics if desired.  However, our general approach is applicable to any vehicle dynamics model.  (For differential drive model, see Appendix \ref{apdx:dyn})





\subsection{Traversability Assessment Models}

\begin{figure}[t!]
    \centering
    \includegraphics[width=\linewidth]{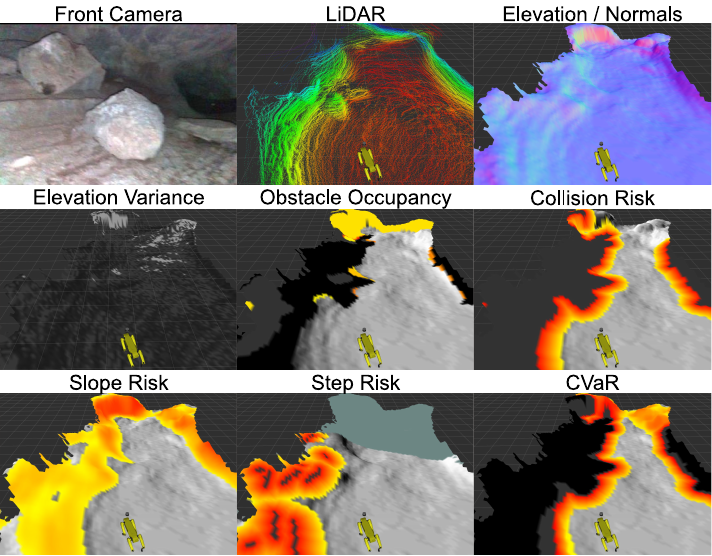}
    \caption{Multi-layer traversability risk analysis, which first aggregates recent pointclouds (top).  Then, each type of analysis (slope, step, collision, etc.) generates a risk map along with uncertainties (middle rows).  These risks are aggregated to compute the final CVaR map (bottom).}
    \label{fig:trav_challenges}
\end{figure}


 The traversability cost is assessed as the combination of multiple risk factors. These factors are designed to capture potential hazards for the target robot in the specific environment (Figure \ref{fig:trav_challenges}). Such factors include:
\begin{itemize}
    \item \textit{Collision}: quantified by the distance to the closest obstacle point.
    \item \textit{Step size}: the height gap between adjacent cells in the grid map. Negative obstacles can also be detected by checking the lack of measurement points in a cell. 
    \item \textit{Tip-over}: a function of slope angles and the robot's orientation. 
    \item \textit{Contact Loss}: insufficient contact with the ground, evaluated by plane-fit residuals.
    \item \textit{Slippage}: quantified by geometry and the surface material of the ground. 
    \item \textit{Sensor Uncertainty}:  sensor and localization error increase the variance of traversability estimates.
\end{itemize}

To efficiently compute the CVaR traversability cost for $l>1$ risk factors, we assume each risk factor $R_l$ is an independent random variable which is normally distributed, with mean $\mu_l$ and variance $\sigma_l$.  We take a weighted average of the risk factors, $R=\sum_lw_lR_l$, which will also be normally distributed as $R\sim\mathcal{N}(\mu,\sigma^2)$. Let $\varphi$ and $\boldsymbol{\Phi}$ denote the probability density function and cumulative distribution function of a standard normal distribution respectively.  The corresponding CVaR is computed as:
\begin{align}
  \rho(R) = \mu + \sigma\frac{\varphi(\boldsymbol{\Phi}^{-1}(\alpha))}{1-\alpha}
  \label{eq:cvar_normal}
\end{align}
We construct $R$ such that the expectation of $R$ is positive, to keep the CVaR value positive.  

Construction of the mean and variance of each risk factor depends on the type of risk.  For example, collision risk is determined by checking for points above the elevation map, and the variance is derived from the elevation map variance, which is mainly a function of localization error.  In contrast, negative obstacle risk is determined by looking for gaps in sensor measurements.  These gaps tend to be a function of sensor sparsity, so the risk variance increases with distance from the sensor frame. 



\subsection{Risk-aware Geometric Planning}

In order to optimize \eqref{eq:optimization_geometric_cost} and \eqref{eq:optimization_geometric_constraint}, the geometric planner computes an optimal path that minimizes the position-dependent dynamic risk metric in \eqref{eq:risk_metric} along the path. Substituting \eqref{eq:cvar_normal} into \eqref{eq:risk_metric}, we obtain:
\begin{align}
    J_{\mathrm{pos}}(x_{0:N})
    &= \mu_0 + \sum_{k=1}^N \bigg[ \mu_k + \sigma_k \frac{\varphi(\boldsymbol{\Phi}^{-1}(\alpha))}{1-\alpha} \bigg]
\end{align}
(For a proof, see Appendix \ref{apdx:cvar}.)

We use the A$^*$ algorithm to solve \eqref{eq:optimization_geometric_cost} over a 2D grid. A$^*$ requires a path cost $g(n)$ and a heuristic cost $h(n)$, given by:
\begin{align}
    g(n) &= J_{\mathrm{pos}}(x_{0:n}) + \lambda \sum_{k=0}^{n-1} \|x_k - x_{k+1}\|^2 \\
    h(n) &= \lambda\,\|x_{n} - x_{G}\|^2
\end{align}
For the heuristic cost we use the shortest Euclidean distance to the goal.  The value of lambda is a relative weighting between the distance penalty and risk penalty and can be thought of as having units of (traversability cost / m).  We use a relatively small value, which means we are mainly concerned with minimizing traversability costs.  


\subsection{Risk-aware Kinodynamic Planning}
\begin{figure}[t!]
    \centering
    \includegraphics[width=\linewidth]{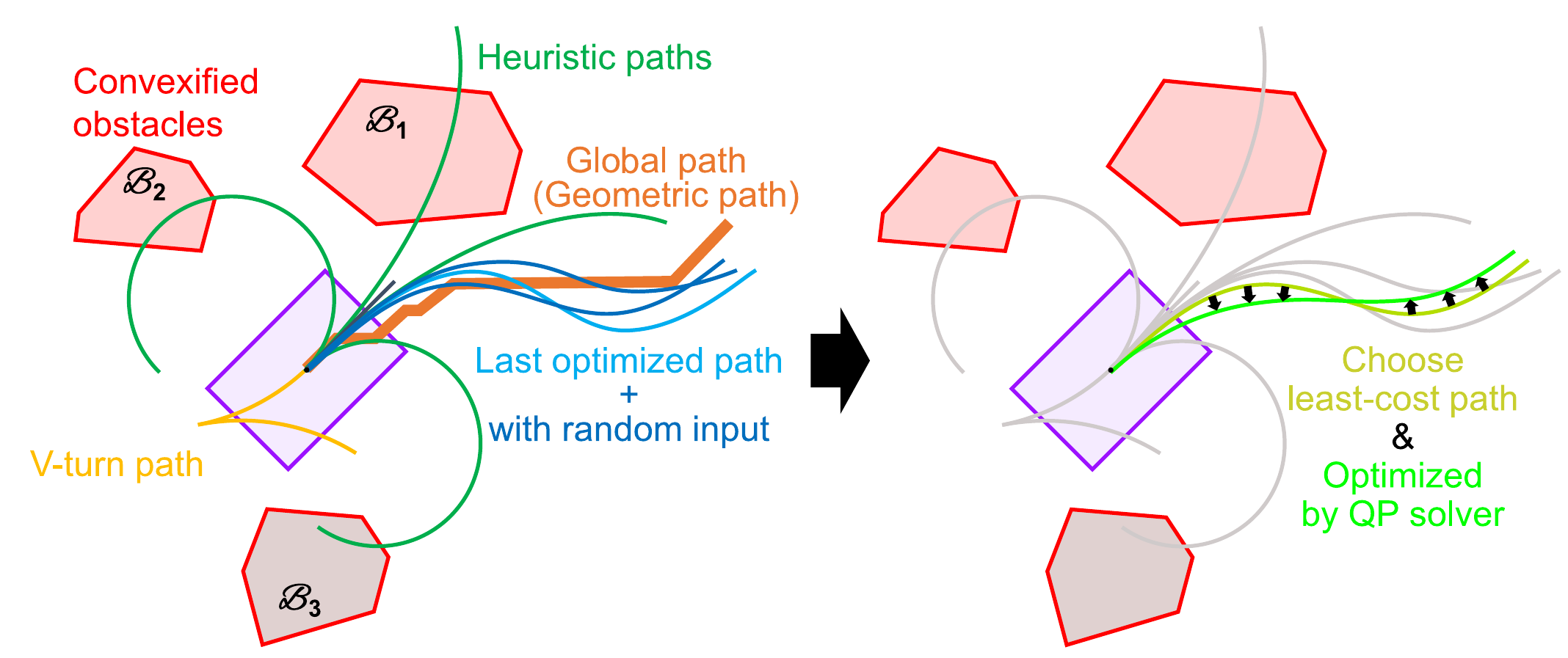}
    \caption{Diagram of kinodynamic MPC planner, which begins with evaluating various paths within a trajectory library.  The lowest cost path is chosen as a candidate and optimized by the QP solver.}
    \label{fig:traj_vis}
\end{figure}

The geometric planner produces a path, i.e. a sequence of poses.  We wish to find a kinodynamically feasible trajectory which stays near this path, while satisfying all constraints and minimizing the CVaR cost.  To solve \eqref{eq:mpc_cost}-\eqref{eq:mpc_state_constraints}, we use a risk-aware kinodynamic MPC planner, whose steps we outline (Figure \ref{fig:traj_vis}, Algorithm \ref{alg:full_algorithm}, Appendix \ref{apdx:traj_fig}).

\begin{algorithm}[t]
    \algsetup{linenosize=\small}
    \small
    \caption{Kinodynamic MPC Planner (sequences $\{\text{var}_k\}_{k=0:T}$ are expressed as $\{\text{var}\}$ for brevity)}
    \begin{algorithmic}[1]
        \renewcommand{\algorithmicrequire}{\textbf{Input:}}
        \renewcommand{\algorithmicensure}{\textbf{Output:}}
        \REQUIRE current state $x_0$, current control sequence (previous solution) $\{u^*\}^{(j)}$
        \ENSURE  re-planned trajectory $\{x^*\}^{(j+1)}$, re-planned control sequence $\{u^*\}^{(j+1)}$
        \\ \textit{Initialization}
        \STATE $\{x^\mathrm{r}\}=$ updateReferenceTrajectory()
        \STATE $\{u^*\}^{(j)}=$ stepControlSequenceForward($\{u^*\}^{(j)}$)
        \\ \textit{Loop process}
        \FOR {$i = 0$ to $\text{qp\_iterations}$}
        \STATE $l$ = generateTrajectoryLibrary($x_0$)
        \STATE $[\{x^\mathrm{c}\}, \{u^\mathrm{c}\}]=$ chooseCandidateFromLibrary($l$)
        \STATE $[\{\delta x^*\}, \{\delta u^*\}]=$ solveQP($\{x^\mathrm{c}\}, \{u^\mathrm{c}\}, \{x^\mathrm{r}\}$)
        \STATE $[\gamma, solved] =$ lineSearch($\{x^\mathrm{c}\}, \{\delta x^*\}, \{u^\mathrm{c}\}, \{\delta u^*\}$)
        \STATE ${u^c_k}=u^\mathrm{c}_k + \gamma \delta u^*_k, ~\forall k=0:T$
        \STATE $\{x^c\}=$ rollOutTrajectory($x_0, \{u^c\}$) 
        \ENDFOR
        \IF {solved}
        \STATE $\{x^*\}^{(j+1)}, \{u^*\}^{(j+1)}=\{x^c\},\{u^c\}$
        \ELSE
        \STATE $\{x^*\}^{(j+1)}, \{u^*\}^{(j+1)}=$ getStoppingTrajectory()
        \ENDIF
        \RETURN $\{x^*\}^{(j+1)}, \{u^*\}^{(j+1)}$
    \end{algorithmic} 
    \label{alg:full_algorithm}
\end{algorithm}






\textit{Trajectory library:} Our kinodynamic planner begins with selecting the best candidate trajectory from a trajectory library, which stores multiple initial control and state sequences.  The selected trajectory is used as initial solution for solving a full optimization problem.  The trajectory library can include:  1) the trajectory accepted in the previous planning iteration, 2) a stopping (braking) trajectory, 3) a geometric plan following trajectory, 4) heuristically defined trajectories (including v-turns, u-turns, and varying curvatures), and 5) randomly perturbed control input sequences.

\textit{QP Optimization:} Next, we construct a non-linear optimization problem with appropriate costs and constraints (\ref{eq:mpc_cost}--\ref{eq:mpc_state_constraints}).  We linearize the problem about the initial solution and solve iteratively in a sequential quadratic programming (SQP) fashion~\cite{nocedal2006numerical}.  Let $\{\hat{x}_k, \hat{u}_k\}_{k=0:T}$ denote an initial solution.  Let $\{\delta x_k, \delta u_k\}_{k=0:T}$ denote deviation from the initial solution.  We introduce the solution vector variable $X$:
\begin{equation}
    X = 
    \begin{bmatrix}
        \delta x_0^\Tr & \cdots & \delta x_T^\Tr & \delta u_0^\Tr & \cdots & \delta u_T^\Tr\\
    \end{bmatrix}^\Tr
\end{equation}
We can then write (\ref{eq:mpc_linear_start}--\ref{eq:mpc_linear_end}) in the form:
\begin{align}
    \text{minimize}\quad & \frac{1}{2}X^\Tr P X + q^\Tr X
    \label{eq:osqp_cost}\\
    \text{subject to}\quad & l\le AX \le u
    \label{eq:osqp_constraints}
\end{align}
where $P$ is a positive semi-definite weight matrix, $q$ is a vector to define the first order term in the objective function, $A$ defines inequality constraints and $l$ and $u$ provide their lower and upper limit. (See Appendix \ref{apdx:qp}.)
In the next subsection we describe these costs and constraints in detail.  This is a quadratic program, which can be solved using commonly available QP solvers.  In our implementation we use the OSQP solver, which is a robust and highly efficient general-purpose solver for convex QPs \cite{osqp}.

\textit{Linesearch:} The solution to the SQP problem returns an optimized variation of the control sequence $\{\delta u_k^*\}_{k=0:T}$.  We then use a linesearch procedure to determine the amount of deviation $\gamma>0$ to add to the current candidate control policy $\pi$: $u_k=u_k + \gamma \delta u^*_k$. (See Appendix \ref{apdx:linesearch}.)

\textit{Stopping Sequence:} If no good solution is found from the linesearch, we pick the lowest cost trajectory from the trajectory library with no collisions.  If all trajectories are in collision, we generate an emergency stopping sequence to slow the robot as much as possible (a collision may occur, but hopefully with minimal energy).

\textit{Tracking Controller:} Having found a feasible and CVaR-minimizing trajectory, we send it to a tracking controller to generate closed-loop tracking behavior at a high rate (>100Hz), which is specific to the robot type (e.g. a simple cascaded PID, or legged locomotive controller).

\subsection{Optimization Costs and Constraints}
\textit{Costs:} Note that \eqref{eq:mpc_cost} contains the CVaR risk.  To linearize this and add it to the QP matrices, we compute the Jacobian and Hessian of $\rho$ with respect to the state $x$.  We efficiently approximate this via numerical differentiation.

\textit{Kinodynamic constraints:} Similar to the cost, we linearize \eqref{eq:mpc_dynamics} with respect to $x$ and $u$.  Depending on the dynamics model, this may be done analytically.

\textit{Control limits:}  We construct the function $g(u)$ in \eqref{eq:mpc_control_constraint} to limit the range of the control inputs.  For example in the 6-state dynamics case, we limit maximum accelerations: $|a_x| < a^{\max}_x$, $|a_y| < a^{\max}_y$, and $|a_{\theta}| < a^{\max}_{\theta}$.

\textit{State limits:}  Within $h(m, x)$ in \eqref{eq:mpc_state_constraints}, we encode velocity constraints: $|v_x| < v_{x}^{\max}$, $|v_y| < v_{y}^{\max}$, and $ |v_\theta| < v_{\theta}^{\max}$.  We also constrain the velocity of the vehicle to be less than some scalar multiple of the risk in that region, along with maximum allowable velocities:
\begin{align}
        |v_\theta| &< \gamma_\theta\,\rho(R_{k}) \\
        \sqrt{v_x^2 + v_y^2} &<  \gamma_v\,\rho(R_{k})
\end{align}
This reduces the energy of interactions the robot has with its environment in riskier situations, preventing more serious damage.

\begin{figure}[t!]
    \centering
    \includegraphics[width=\linewidth]{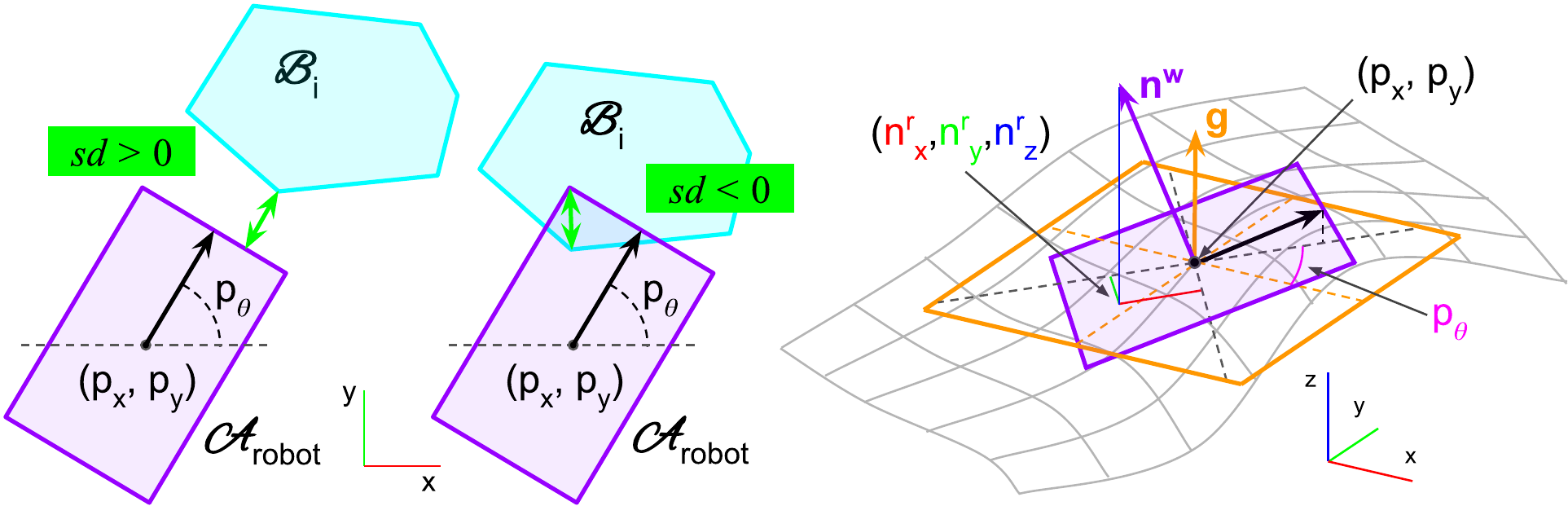}
    \caption{Left: Computing convex to convex signed distance function between the robot footprint and an obstacle.  Signed distance is positive with no intersection and negative with intersection.  Right: Robot pitch and roll are computed from the surface normal rotated by the yaw of the robot.  Purple rectangle is the robot footprint with surface normal $n^w$.  $\bm{g}$ denotes gravity vector, $n^r_{x,y,z}$ are the robot-centric surface normal components used for computing pitch and roll.}
    \label{fig:slope_angle}
\end{figure}

\textit{Position risk constraints:}  Within $h(m, x_k)$ we would like to add constraints on position and orientation to prevent the robot from hitting obstacles.  The general form of this constraint is:
\begin{align}
    \rho(R_k) < \rho^{\max}
\end{align}
To create this constraint, we locate areas on the map where the risk $\rho$ is greater than the maximum allowable risk.  These areas are marked as obstacles, and are highly non-convex.  To obtain a convex and tractable approximation of this highly non-convex constraint, we decompose obstacles into non-overlapping 2D convex polygons, and create a signed distance function which determines the minimum distance between the robot's footprint (also a convex polygon) and each obstacle \cite{schulman2014motion}.  Let $\mathcal{A}, \mathcal{B}\subset \mathbb{R}^2$ be two sets, and define the distance between them as:
\begin{align}
    \mathrm{dist}(\mathcal{A}, \mathcal{B}) = \inf\{\|T\| ~ | ~ (T+\mathcal{A})\cap\mathcal{B} \neq \emptyset\}
\end{align}
where $T$ is a translation.  When the two sets are overlapping, define the penetration distance as:
\begin{align}
    \mathrm{penetration}(\mathcal{A}, \mathcal{B}) = \inf\{\|T\| ~ | ~ (T+\mathcal{A})\cap\mathcal{B} = \emptyset\}
\end{align}
Then we can define the signed distance between the two sets as:
\begin{align}
    \mathrm{sd}(\mathcal{A}, \mathcal{B})=\mathrm{dist}(\mathcal{A},\mathcal{B}) - \mathrm{penetration}(\mathcal{A},\mathcal{B})
\end{align}
We then include within $h(m,x_k)$ a constraint to enforce the following inequality:
\begin{align}
    \mathrm{sd}(\mathcal{A}_{\mathrm{robot}},\mathcal{B}_i) > 0 \quad \forall i \in \{0,\cdots,N_{\mathrm{obstacles}}\}
\end{align}
Note that the robot footprint $\mathcal{A}_{\mathrm{robot}}$ depends on the current robot position and orientation: $\mathcal{A}_{\mathrm{robot}}(p_x, p_y, p_{\theta})$, while each obstacle $\mathcal{B}_i(m)$ is dependent on the information in the map (See Figure \ref{fig:slope_angle}).

\textit{Orientation constraints:}
We wish to constrain the robot's orientation on sloped terrain in such a way as to prevent the robot from rolling over or performing dangerous maneuvers.  To do this, we add constraints to $h(m, x_k)$ which limit the roll and pitch of the robot as it settles on the surface of the ground.  Denote the position as $p=[p_x,p_y]^\intercal$ and the position/yaw as $s=[p_x,p_y,p_\theta]^{\intercal}$.  Let the robot's pitch be $\psi$ and roll be $\phi$ in its body frame.  Let $\omega=[\psi, \phi]^\intercal$.  The constraint will have the form $|\omega| \prec \omega^{\max}$.  At $p$, we compute the surface normal vector, call it $n^w=[n^w_x,n^w_y,n^w_z]^\intercal$, in the world frame.  Let $n^r = [n^r_x,n^r_y,n^r_z]^\intercal$, be the surface normal in the body frame, where we rotate by the robot's yaw: $n^r = R_\theta n^w$ (see Figure \ref{fig:slope_angle}), where $R_\theta$ is a basic rotation matrix by the angle $\theta$ about the world $z$ axis.
Then, we define the robot pitch and roll as $\omega = g(n^r)$ where:
\begin{align}
\omega = g(n^r) =
\begin{bmatrix}
    \mathrm{atan2}(n^r_x,n^r_z)\\
    -\mathrm{atan2}(n^r_y,n^r_z)
\end{bmatrix}
\end{align}
Note that $\omega$ is a function of $s$.  Creating a linearly-constrained problem requires a linear approximation of the constraint: 
\begin{equation}
    \label{eq:orient_con}
    |\nabla_s\omega(s)\delta s + \omega(s)| < \omega^{\max}
\end{equation}
This is accomplished by finding the gradients with respect to position and yaw separately (See Appendix \ref{apdx:grad}).

\textit{Box Constraint:} Note that if $\delta x$ and $\delta u$ are too large, linearization errors will dominate.  To mitigate this we also include box constraints within \eqref{eq:mpc_control_constraint} and \eqref{eq:mpc_state_constraints} to maintain a bounded deviation from the initial solution: $|\delta x| < \epsilon_x$ and $|\delta u| < \epsilon_u$.

\textit{Adding Slack Variables:}  To further improve the feasibility of the optimization problem we introduce auxilliary slack variables for constraints on state limits, position risk, and orientation.  For a given constraint $h(x)>0$ we introduce the slack variable $\epsilon$, and modify the constraint to be $h(x)>\epsilon$ and $\epsilon<0$.  We then penalize large slack variables with a quadratic cost: $\lambda_\epsilon \epsilon^2$.  These are incorporated into the QP problem \eqref{eq:osqp_cost} and \eqref{eq:osqp_constraints}.


\subsection{Dynamic Risk Adjustment}

The CVaR metrics allows us to dynamically adjust the level and severity of risk we are willing to accept. Selecting low $\alpha$ reverts towards using the mean cost as a metric, leading to optimistic decision making while ignoring low-probability but high cost events. Conversely, selecting a high $\alpha$ leans towards conservatism, reducing the likelihood of fatal events while reducing the set of possible paths. We adjust $\alpha$ according to two criteria:  1)  Mission-level states, where depending on the robot's role, or the balance of environment and robot capabilities, the risk posture for individual robots may differ.  2)  Recovery Behaviors, where if the robot is trapped in an unfavorable condition, by gradually decreasing $\alpha$, an escape plan can be found with minimal risk.  These heuristics are especially useful in the case of risk-aware planning, because the feasibility of online nonlinear MPC is difficult to guarantee.  When no feasible solution is found for a given risk level $\alpha$, a riskier but feasible solution can be quickly found and executed.


\section{Experiments}

In this section, we report the performance of STEP. We first present a comparative study between different adjustable risk thresholds in simulation on a wheeled differential drive platform.  Then, we demonstrate real-world performance using a wheeled robot deployed in an abandoned subway filled with clutter, and a legged platform deployed in a lava tube environment.

\subsection{Simulation Study}

\begin{figure}
    \centering
    \includegraphics[width=0.49\textwidth]{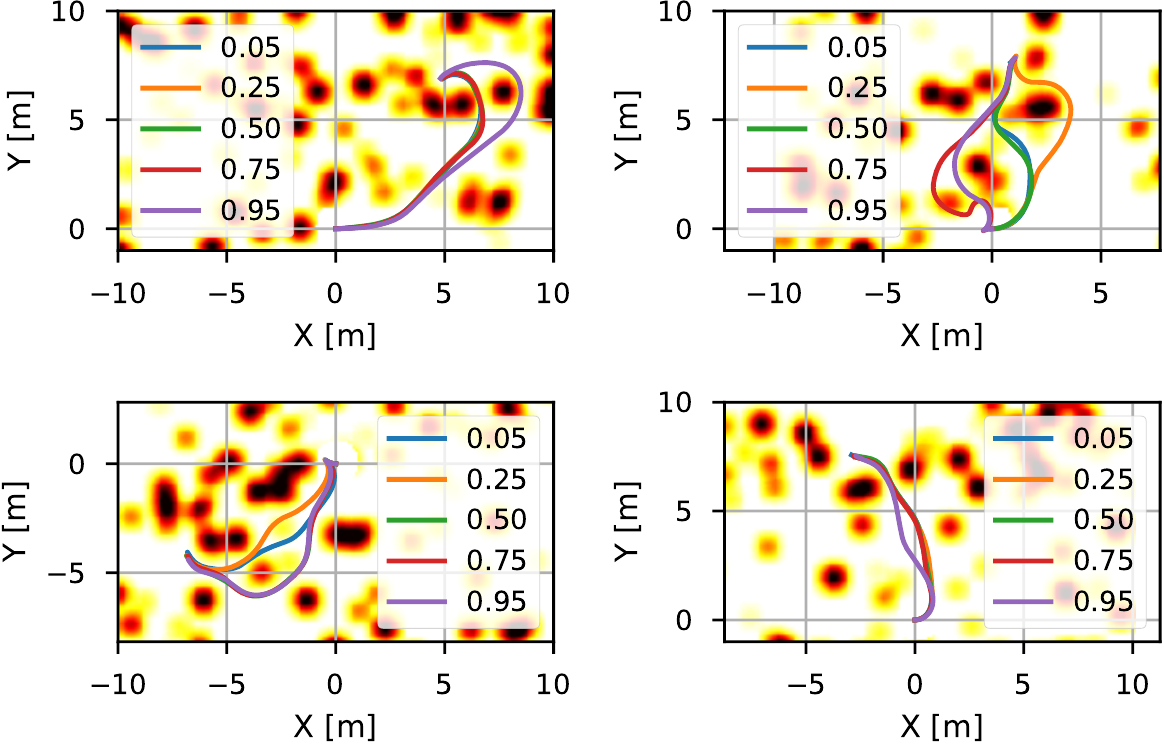}
    \caption{Path distributions from four simulated runs. The risk level $\alpha$ spans from 0.1 (close to mean-value) to 0.95 (conservative). Smaller $\alpha$ typically results in a shorter path, while larger $\alpha$ chooses statistically safe paths.}
    \label{fig:sim_pathdist}
\end{figure}

\begin{figure}
    \centering
    \includegraphics[width=0.49\textwidth]{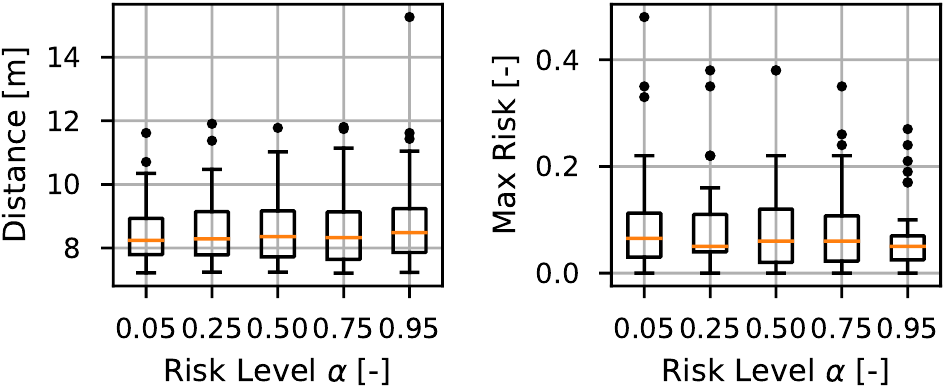}
    \caption{Distance vs risk trade-off from 50 Monte-Carlo simulations. Left: Distributions of path distance. Right: Distributions of max risk along the traversed paths.  Box plot uses standard quartile format and dots are outliers.}
    \label{fig:sim_tradeoff}
\end{figure}

To assess statistical performance, we perform 50 Monte-Carlo simulations with randomly generated maps and goals. Random traversability costs are assigned to each grid cell. The following assumptions are made: 1) no localization error, 2) no tracking error, and 3) a simplified perception model with artificial noise. We give a random goal 8\,m away and evaluate the path cost and distance.  We use a differential-drive dynamics model (no lateral velocity).

We compare STEP using different $\alpha$ levels.  Figure \ref{fig:sim_pathdist} shows the distribution of paths for different planning configurations. The optimistic (close to mean-value) planner $\alpha=0.05$ 
typically generates shorter paths, while the conservative setting $\alpha=0.95$ makes long detours to select statistically safer paths. The other $\alpha$ settings show distributions between these two extremes, with larger $\alpha$ generating similar paths to the conservative planner and smaller $\alpha$ generating more time-optimal paths. Statistics are shown in Figure \ref{fig:sim_tradeoff}.

\subsection{Hardware Results}
\begin{figure*}[ht]
    \centering
    \includegraphics[width=\textwidth]{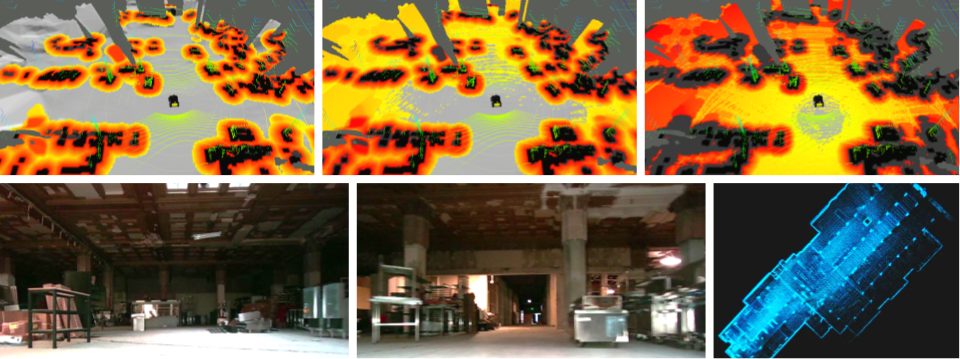}
    \caption{Traversability analysis results for Husky in an abandoned subway experiment. Top left to right: Risk maps at three varying risk levels: $\alpha = 0.1, ~0.5, ~0.9$, respectively. Colors correspond to CVaR value (white: safe ($r<=0.05$), yellow to red: moderate ($0.05 < r <= 0.5$), black: risky ($r > 0.5$)).  Also shown are the most recent LiDAR measurements (green points).  Bottom left and middle:  Front and right on-board cameras observing the same location.  Bottom right:  Completed top-down map of the environment after autonomous exploration.  Bright dots are pillars, which are visible in the camera images.  }
    \label{fig:husky_subway}
\end{figure*}

We deployed STEP on two different robots (wheeled and legged) in two different challenging environments (an abandoned subway and a lava tube).  First we tested STEP on a Clearpath Husky robot in an abandoned subway filled with industrial clutter.  The robot was equipped with custom sensing and computing units, and driven by JPL's NeBula autonomy software \cite{AliNeBula21}.  3 Velodyne VLP-16s were used for collecting LiDAR data.  Localization was provided onboard by a LIDAR-based SLAM solution \cite{Palieri2020,Ebadi2020}.  The entire autonomy stack runs on an Intel Core i7 CPU. The typical CPU usage for the traversability stack is about a single core.  The robot successfully explored two levels of the approximately 200m x 100m environment.  In Figure \ref{fig:husky_subway}, we plot the risk map in this environment at varying levels of $\alpha$.  We clearly see the effects on the risk map, where higher values of $\alpha$ result in closing narrow openings, assigning high cost to unknown regions, and increasing the size of obstacles.  The effect of these risk analyses results in intuitive outcomes - for example, a low pile of metal, while probably traversable, should be avoided if possible.  When the region has inadequate sensor coverage, the risk will be high.  When the robot is closer and the sensor coverage is good, then the CVaR cost will decrease, yielding a more accurate risk assessment.  This results in more efficient and safer planning when compared to deterministic methods.  For example, in our prior work \cite{thakker2021autonomous}, the deterministic approach led to frequent oscillations in planning as obstacles appeared and disappeared with sensor and localization noise.

Next, we deployed STEP on a Boston Dynamics Spot quadruped robot at the Valentine Cave in Lava Beds National Monument, Tulelake, CA. 
The main sensor for localization and traversability analysis is a Velodyne VLP-16, fused with Spot's internal Intel realsense data to cover blind spots.  The payload was similar to that of Husky, proving on-board, real-time computing.

\begin{figure*}[ht]
    \centering
    \includegraphics[width=\textwidth]{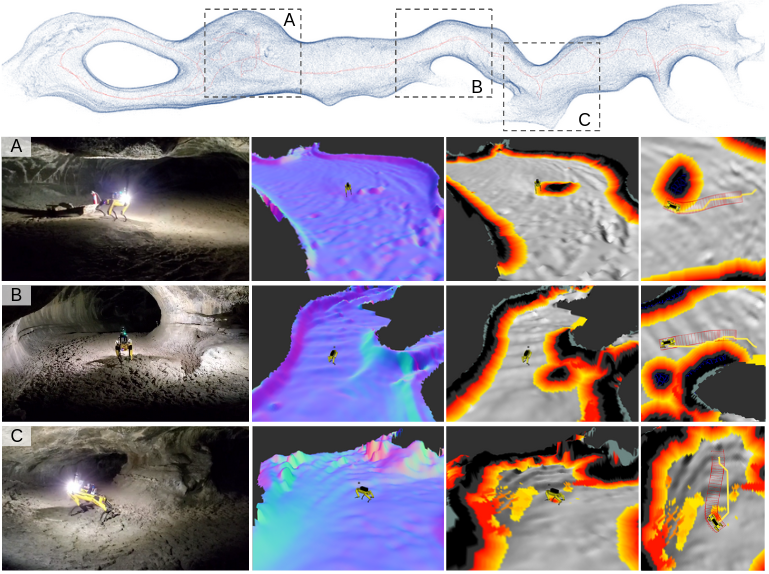}
    \caption{Traversability analysis results for the Valentine Cave experiment. From left to right: Third-person view, elevation map (colored by normal direction), risk map (colored by risk level. white: safe ($r<=0.05$), yellow to red: moderate ($0.05 < r <= 0.5$), black: risky ($r > 0.5$)), and planned geometric/kinodynamic paths (yellow lines/red boxes). }
    \label{fig:valentine_result}
\end{figure*}

Figure \ref{fig:valentine_result} shows the interior of the cave and algorithm's representations. The rough ground surface, rounded walls, ancient lava waterfalls, steep non-uniform slopes, and boulders all pose significant traversability stresses. Furthermore, there are many occluded places which affect the confidence in traversability estimates.

We tested our risk-aware traversability software during our fully autonomous runs. The planner was able to navigate the robot safely to the every goal provided by the upper-layer coverage planner \cite{Bouman2020,kim2021plgrim} despite the challenges posed by the environment. Figure \ref{fig:valentine_result} shows snapshots of elevation maps, CVaR risk maps, and planned paths. The risk map captures walls, rocks, high slopes, and ground roughness as mobility risks. STEP enables Spot to safely traverse the entire extent of the lava tube, fully exploring all regions.  STEP navigates 420 meters over 24 minutes, covering 1205 square meters of rough terrain.

\section{Conclusion}
We have presented STEP (Stochastic Traversability Evaluation and Planning), our approach for autonomous robotic navigation in unsafe, unstructured, and unknown environments.  We believe this approach finds a sweet-spot between computation, resiliency, performance, and flexibility when compared to other motion planning approaches in such extreme environments.  Our method is generalizable and extensible to a wide range of robot types, sizes, and speeds, as well as a wide range of environments.  Our future work includes robustification of subcomponents and extension to higher speeds.

\section*{Acknowledgement}
The work is performed at the Jet Propulsion Laboratory, California Institute of Technology, under a contract with the National Aeronautics and Space Administration (80NM0018D0004), and Defense Advanced Research Projects Agency (DARPA).

\printbibliography

@inproceedings{kalita2018path,
  title={Path planning and navigation inside off-world lava tubes and caves},
  author={Kalita, Himangshu and Morad, Steven and Ravindran, Aaditya and Thangavelautham, Jekan},
  booktitle={IEEE/ION Position, Location and Navigation Symposium},
  pages={1311-1318},
  year={2018}
}

@article{osqp,
  author  = {Stellato, B. and Banjac, G. and Goulart, P. and Bemporad, A. and Boyd, S.},
  title   = {{OSQP}: an operator splitting solver for quadratic programs},
  journal = {Mathematical Programming Computation},
  year    = {2020},
  pages   = {1-36}
}

@article{schulman2014motion,
  title={Motion planning with sequential convex optimization and convex collision checking},
  author={Schulman, John and Duan, Yan and Ho, Jonathan and Lee, Alex and Awwal, Ibrahim and Bradlow, Henry and Pan, Jia and Patil, Sachin and Goldberg, Ken and Abbeel, Pieter},
  journal={The International Journal of Robotics Research},
  year={2014},
  volume={33},
  number={9},
  pages={1251-1270},
  publisher={SAGE Publications Sage UK: London, England}
}

@article{agha2017CRM,
  title={Confidence-rich grid mapping},
  author={Agha-Mohammadi, A. and Heiden, Eric and Hausman, Karol and Sukhatme, Gaurav},
  journal={The International Journal of Robotics Research},
  year={2017},
  volume={38},
  number={12-13},
  pages={1352-1374}
}

@article{singh2018framework,
      title={A Framework for Time-Consistent, Risk-Sensitive Model Predictive Control: Theory and Algorithms}, 
      author={Singh, Sumeet and Chow, Y. and Majumdar, Anirudha and Marco Pavone},
      year={2018},
      journal={IEEE Transactions on Automatic Control},
      volume={64},
      number={7},
      pages={2905-2912}
}

@article{ruszczynski2014riskaverseDP,
    author = {Ruszczynski, Andrzej},
    year = {2014},
    title = {Risk-averse dynamic programming for Markov decision processes},
    journal = {Mathematical Programming},
    volume={75},
    number={2},
    pages={235-261}
}

@book{nocedal2006numerical,
  title={Numerical optimization},
  author={Nocedal, Jorge and Wright, Stephen},
  year={2006},
  publisher={Springer Science \& Business Media}
}

@article{ono2015chance,
  title={Chance-constrained dynamic programming with application to risk-aware robotic space exploration},
  author={Ono, Masahiro and Pavone, Marco and Kuwata, Kuwata and Balaram, J. (Bob)},
  journal={Autonomous Robots},
  year={2015},
  volume={39},
  number={4},
  pages={555-571}
}

@article{wang2020non,
  title={Non-gaussian chance-constrained trajectory planning for autonomous vehicles under agent uncertainty},
  author={Allen Wang and Ashkan Jasour and Brian Williams},
  journal={Robotics and Automation Letters},
  year={2020},
  publisher={IEEE},
  volume={5},
  number={4},
  pages={6041-6048}
}

@inproceedings{koenig1994risk,
  title={Risk-sensitive planning with probabilistic decision graphs},
  author={Koenig, Sven and Simmons, Reid G},
  booktitle={Principles of Knowledge Representation and Reasoning},
  year={1994},
  pages={363-373},
  organization={Elsevier}
}

@inproceedings{xu2010distributionally,
  title={{Distributionally robust Markov decision processes}},
  author={Xu, Huan and Mannor, Shie},
  booktitle={Advances in Neural Information Processing Systems},
  pages={2505-2513},
  year={2010}
}

@incollection{majumdar2020should,
  title={How should a robot assess risk? Towards an axiomatic theory of risk in robotics},
  author={Anirudha Majumdar and Marco Pavone},
  booktitle={Robotics Research},
  year={2020},
  pages={75-84},
  publisher={Springer}
}

@article{artzner1999coherent,
  title={Coherent measures of risk},
  author={Artzner, Philippe and Delbaen, Freddy and Eber, J. and Heath, David},
  journal={Mathematical finance},
  volume={9},
  number={3},
  pages={203--228},
  year={1999},
  publisher={Wiley Online Library}
}

@inproceedings{chow2015risk,
  title={{Risk-sensitive and robust decision-making: a CVaR optimization approach}},
  author={Chow, Yinlam and Tamar, Aviv and Mannor, Shie and Pavone, Marco},
  booktitle={Advances in Neural Information Processing Systems},
  pages={1522-1530},
  year={2015}
}

@book{camacho2013model,
  title={Model predictive control},
  author={Camacho, Eduardo F and Alba, Carlos Bordons},
  year={2013},
  publisher={Springer Science \& Business Media}
}

@inproceedings{kalakrishnan2011stomp,
  title={{STOMP: Stochastic trajectory optimization for motion planning}},
  author={Kalakrishnan, Mrinal and Chitta, Sachin and Theodorou, Evangelos and Pastor, Peter and Schaal, Stefan},
  booktitle={IEEE International Conference on Robotics and Automation},
  year={2011},
  pages={4569-4574}
}

@article{boggs1995sequential,
  title={Sequential quadratic programming},
  author={Boggs, Paul T and Tolle, Jon W},
  journal={Acta numerica},
  pages={529-562},
  year={1995}
}

@inproceedings{ghosh2018pace,
  author={S. {Ghosh} and K. {Otsu} and M. {Ono}},
  booktitle={IEEE/RSJ International Conference on Intelligent Robots and Systems}, 
  title={Probabilistic Kinematic State Estimation for Motion Planning of Planetary Rovers}, 
  year={2018},
  pages={5148-5154}
 }

@article{hait2002algorithms,
author = { Alain   Ha{\"i}t  and  Thierry   Simeon  and  Michel   Ta{\"i}x },
title = {Algorithms for rough terrain trajectory planning},
journal = {Advanced Robotics},
year  = {2002},
volume = {16},
number ={8},
pages = {673-699},
publisher = {Taylor & Francis}}

@article{hakobyan2019risk,
  title={{Risk-aware motion planning and control using CVaR-constrained optimization}},
  author={Hakobyan, Astghik and Kim, Gyeong Chan and Yang, Insoon},
  journal={IEEE Robotics and Automation Letters},
  year={2019},
  volume={4},
  number={4},
  pages={3924-3931},
  publisher={IEEE}
}

@article{papdakis2013survey,
title = "Terrain traversability analysis methods for unmanned ground vehicles: A survey",
journal = "Engineering Applications of Artificial Intelligence",
year = "2013",
volume={26},
number={4},
pages={1373-1385},
author = "Panagiotis Papadakis",
}

@article{fankhauser2018probabilisticterrain,
  author={P{\'{e}}ter {Fankhauser} and Michael {Bloesch} and Marco {Hutter}},
  journal={IEEE Robotics and Automation Letters}, 
  title={Probabilistic Terrain Mapping for Mobile Robots With Uncertain Localization}, 
  year={2018},
  volume={3},
  number={4},
  pages={3019-3026}
}

@article{Fankhauser2016GridMapLibrary,
  author = {Fankhauser, P{\'{e}}ter and Hutter, Marco},
  journal = {Robot Operating System (ROS), The Complete Reference},
  title = {A Universal Grid Map Library: Implementation and Use Case for Rough Terrain Navigation},
  pages = {99-120},
  year = {2016}
}

@article{thakker2021autonomous,
  title={Autonomous off-road navigation over extreme terrains with perceptually-challenging conditions},
  author={Thakker, Rohan and Alatur, Nikhilesh and Fan, David D and Tordesillas, Jesus and Paton, Michael and Otsu, Kyohei and Toupet, Olivier and Agha-mohammadi, A.},
  journal={International Symposium on Experimental Robotics}, 
  year={2020}
}

@article{Bouman2020,
  title={{Autonomous Spot: Long-range autonomous exploration of extreme environments with legged locomotion}},
  author={Bouman, Amanda and Ginting, Muhammad Fadhil and Alatur, Nikhilesh and Palieri, Matteo and Fan, David D and Touma, Thomas and Pailevanian, Torkom and Kim, S. and Otsu, Kyohei and Burdick, Joel and Agha-mohammadi, A.},
  journal={IEEE/RSJ International Conference on Intelligent Robots and Systems}, 
  year={2020}
}

@article{rockafellar2000optimization,
  title={Optimization of conditional value-at-risk},
  author={Rockafellar, R Tyrrell and Uryasev, Stanislav and others},
  journal={Journal of Risk},
  volume={2},
  number={3},
  pages={21-41},
  year={2000}
}

@INPROCEEDINGS{dixit2020risksensitive,
      title={Risk-Sensitive Motion Planning using Entropic Value-at-Risk}, 
      author={Anushri Dixit and Mohamadreza Ahmadi and Joel W. Burdick},
      year={2021},
      booktitle={European Control Conference}
}

@article{otsu2020fast,
  title={Fast approximate clearance evaluation for rovers with articulated suspension systems},
  author={Otsu, Kyohei and Matheron, Guillaume and Ghosh, Sourish and Toupet, Olivier and Ono, Masahiro},
  journal={Journal of Field Robotics},
  year={2020},
  volume={37},
  issue={5},
  pages={768-785},
  publisher={Wiley Online Library}
}

@INPROCEEDINGS{ahmadi2020uncertainty,
  author={Mohamadreza {Ahmadi} and Masahiro {Ono} and Michel D. {Ingham} and Richard M. {Murray} and Aaron D. {Ames}},
  booktitle={American Control Conference}, 
  title={Risk-Averse Planning Under Uncertainty}, 
  year={2020},
  volume={},
  number={},
  pages={3305-3312}}

@inproceedings{ahmadi2020constrained,
  author={Mohamadreza Ahmadi and Ugo Rosolia and Michel D. Ingham and Richard M. Murray and Aaron D. Ames},
  booktitle={AAAI Conference on Artificial Intelligence}, 
  title={Constrained Risk-Averse Markov Decision Processes},
  volume={},
  number={},
  pages={},
  year={2021}
}

@inproceedings{gestalt,
  title={Stereo vision and rover navigation software for planetary exploration},
  author={Goldberg, Steven B and Maimone, Mark W and Matthies, Larry},
  booktitle={IEEE Aerospace Conference},
  year={2002},
  organization={IEEE}
}

@inproceedings{lew2020chance,
  title={Chance-constrained sequential convex programming for robust trajectory optimization},
  author={Lew, Thomas and Bonalli, Riccardo and Pavone, Marco},
  booktitle={2020 European Control Conference (ECC)},
  pages={1871--1878},
  year={2020},
  organization={IEEE}
}

@article{lofberg2012oops,
  title={Oops! I cannot do it again: Testing for recursive feasibility in MPC},
  author={L{\"o}fberg, Johan},
  journal={Automatica},
  volume={48},
  number={3},
  pages={550--555},
  year={2012},
  publisher={Elsevier}
}

@article{fan2020deep,
  title={Deep learning tubes for Tube MPC},
  author={Fan, David D and Agha-mohammadi, Ali-akbar and Theodorou, Evangelos A},
  journal={Robotics: Science and Systems (RSS)},
  year={2020}
}

@inproceedings{dabney2018distributional,
author = {Dabney, Will and Rowland, Mark and Bellemare, Marc G and Munos, R{\'{e}}mi},
booktitle = {Thirty-Second AAAI Conference on Artificial Intelligence},
title = {{Distributional reinforcement learning with quantile regression}},
year = {2018}
}

@article{AliNeBula21,
  title={{NeBula}: Quest for robotic autonomy in challenging environments; {TEAM} {CoSTAR} at the {DARPA Subterranean Challenge}},
  author={{{A.} Agha-mohammadi, et al.}},
  journal={Field Robotics},
  year={2021},
}

@inproceedings{kim2021plgrim,
  author = {Kim, Sung-Kyun and Bouman, Amanda and Salhotra, Gautam and Fan, David D. and Otsu, Kyohei and Burdick, Joel and Agha-mohammadi, A.},
  title = {{PLGRIM}: Hierarchical value learning for large-scale exploration in unknown environments},
  booktitle={International Conference on Automated Planning and Scheduling (ICAPS)},
  volume={31},
  year={2021}
}

@article{Palieri2020,
  title={{LOCUS}: A multi-sensor lidar-centric solution for high-precision odometry and {3D} mapping in real-time},
  author={Palieri, Matteo and Morrell, Benjamin and Thakur, Abhishek and Ebadi, Kamak and Nash, Jeremy and Chatterjee, Arghya and Kanellakis, Christoforos and Carlone, Luca and Guaragnella, Cataldo and Agha-mohammadi, A.},
  journal={IEEE Robotics and Automation Letters},
  volume={6},
  number={2},
  pages={421--428},
  year={2020},
  publisher={IEEE}
}

@inproceedings{Ebadi2020,
author = {Ebadi, K and Chang, Y and Palieri, M and Stephens, A and Hatteland, A and Heiden, E and Thakur, A and Morrell, B and Carlone, L and Agha-mohammadi, A},
booktitle = {IEEE International Conference on Robotics and Automation},
title = {{LAMP}: Large-scale autonomous mapping and positioning for exploration of perceptually-degraded subterranean environments},
pages = {80--86},
year = {2020}
}

@article{ahmadi2021riskaverse,
      title={Risk-Averse Stochastic Shortest Path Planning}, 
      author={Mohamadreza Ahmadi and Anushri Dixit and Joel W. Burdick and Aaron D. Ames},
      year={2021},
      journal={arXiv:2103.14727},
}

\clearpage
\newpage
\appendix
\subsection{Dynamics model for differential drive}
\label{apdx:dyn}
For a simple system which produces forward velocity and steering, (e.g. differential drive systems), we may wish to specify the state and controls as:
\begin{align}
    x = [p_x, p_y, p_\theta, v_x]^{\intercal} \\
    u = [a_x, v_{\theta}]^{\intercal}
\end{align}
For example, the dynamics $x_{k+1} = f(x_k,u_k)$ for a simple differential-drive system can be written as:
\begin{align}
    x_{k+1} = x_k + \Delta t \begin{bmatrix}
        v_x \cos(p_\theta)\\
        v_x \sin(p_\theta)\\
        \gamma v_x + (1-\gamma) v_{\theta}\\
        a_x
    \end{bmatrix}
\end{align}
where $\gamma\in[0,1]$ is a constant which adjusts the amount of turning-in-place the vehicle is permitted.

\subsection{Computing the dynamic risk metric using CVaR for Normal distributions}
\label{apdx:cvar}
\begin{equation}
\begin{aligned}
    J(x_0, \pi) &= R_0 + \rho_0\big( R_1 + \rho_{1}\big(R_2 + \dotsc + \rho_{T-1}(R_{T})\big)\big) \\
    &= R_0 + \rho\Bigg( R_1 + \rho\Big(R_2 + \dotsc + \rho(R_{T-1} + \\ &\qquad\mu_T + \sigma_T \frac{\varphi(\boldsymbol{\Phi}^{-1}(\alpha))}{1-\alpha})\Big)\Bigg) \\
    &= R_0 + \rho\Bigg( R_1 + \rho\Big(R_2 + \dotsc + \rho(R_{T-2} + \\  &\qquad\mu_{T-1} + \mu_T + (\sigma_{T-1} + \sigma_T) \frac{\varphi(\boldsymbol{\Phi}^{-1}(\alpha))}{1-\alpha})\Big)\Bigg) \\
    & \vdots \\
    &= R_0 + \sum_{i=1}^{T}\Bigg(\mu_i + \sigma_i \frac{\varphi(\boldsymbol{\Phi}^{-1}(\alpha))}{1-\alpha}\Bigg)\\
    &= \sum_{i=0}^{T}\rho(R_i)\nonumber
\end{aligned}
\end{equation}

\subsection{Kinodynamic Planning Diagram}
\label{apdx:traj_fig}

\begin{figure}[h!]
    \centering
    \includegraphics[width=\linewidth]{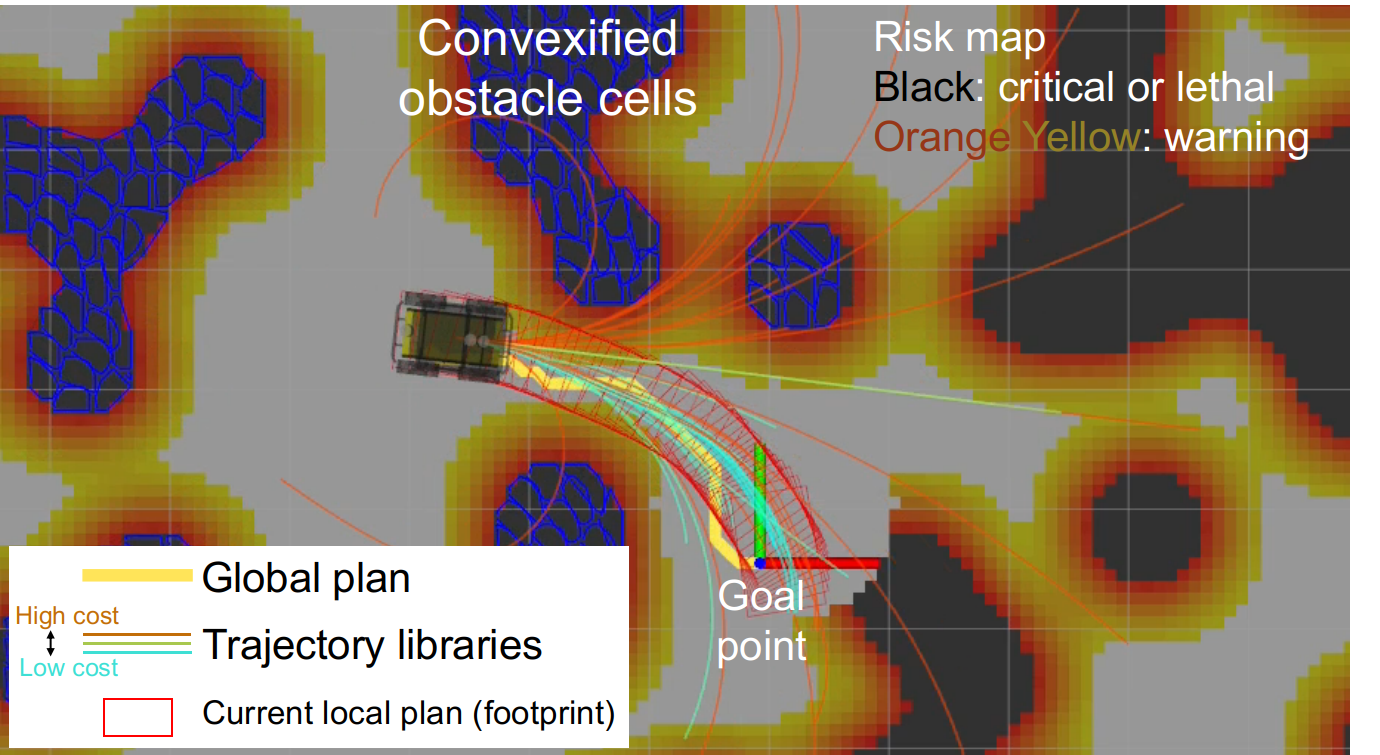}
    \caption{Diagram of kinodynamic MPC planner, which begins with evaluating various paths within a trajectory library.  The lowest cost path is chosen as a candidate and optimized by the QP solver.}
\end{figure}

\subsection{Gradients for Orientation Constraint}
\label{apdx:grad}
We describe in further detail the derivation of the orientation constraints.  Denote the position as $p=[p_x,p_y]^\intercal$ and the position/yaw as $s=[p_x,p_y,p_\theta]$.  We wish to find the robot's pitch $\psi$ and roll $\phi$ in its body frame.  Let $\omega=[\psi, \phi]^\intercal$.  The constraint will have the form $|\omega(s)| <= \omega_{max}$.  At $p$, we compute the surface normal vector, call it $n^w=[n^w_x,n^w_y,n^w_z]^\intercal$, in the world frame.  To convert the normal vector in the body frame, $n^r = [n^r_x,n^r_y,n^r_z]^\intercal$, we rotate by the robot's yaw: $n^r = R_\theta n^w$ (see Figure \ref{fig:slope_angle}), where $R_\theta$ is a basic rotation matrix by the angle $\theta$ about the world $z$ axis:
\begin{equation}
    R_\theta = \begin{bmatrix}
        \cos{p_\theta} && \sin{p_\theta} && 0\\
        -\sin{p_\theta} && \cos{p_\theta} && 0\\
        0 && 0 && 1
    \end{bmatrix}
\end{equation}
Let the robot pitch and roll vector $\omega$ be defined as $\omega = g(n^r)$, where:
\begin{align}
\omega = g(n^r) =
\begin{bmatrix}
    \mathrm{atan2}(n^r_x,n^r_z)\\
    -\mathrm{atan2}(n^r_y,n^r_z)
\end{bmatrix}
\end{align}
Creating a linearly-constrained problem requires a linear approximation of the constraint: 
\begin{equation}
    |\nabla_s\omega(s)\delta s + \omega(s)| <= \omega_{max}
\end{equation}
Conveniently, computing $\nabla_s\omega(s)$ reduces to finding gradients w.r.t position and yaw separately.  Let $\nabla_s\omega(s) = [\nabla_p\omega(s), \nabla_\theta\omega(s)]^\intercal$, then:
\begin{align}
    \nabla_p\omega(s) &= (\nabla_{n^r}g)(R_\theta)(\nabla_{p} n^w) \\
    \nabla_\theta\omega(s) &= (\nabla_{n^r}g)(\frac{d}{d\theta}R_\theta)(n^w)
\end{align}
where:
\begin{equation}
    \nabla_{n^r}g = \begin{bmatrix}
        \frac{n^r_z}{(n^r_x)^2+(n^r_z)^2} && 0 && \frac{-n^r_x}{(n^r_x)^2+(n^r_z)^2} \\
        0 && \frac{-n^r_z}{(n^r_y)^2+(n^r_z)^2} && \frac{n^r_y}{(n^r_y)^2+(n^r_z)^2}
    \end{bmatrix}
\end{equation}
and 
\begin{equation}
    \nabla_p n^w = \begin{bmatrix}
        \frac{\partial n^w_x}{\partial p_x} && \frac{\partial n^w_x}{\partial p_y} \\
        \frac{\partial n^w_y}{\partial p_x} && \frac{\partial n^w_y}{\partial p_y} \\
        \frac{\partial n^w_z}{\partial p_x} && \frac{\partial n^w_z}{\partial p_y} \\
    \end{bmatrix}
\end{equation}
The terms with the form $\frac{\partial n^w_x}{\partial p_x}$ amount to computing a second-order gradient of the elevation on the 2.5D map.  This can be done efficiently with numerical methods \cite{Fankhauser2016GridMapLibrary}.

\subsection{Converting non-linear MPC problem to a QP problem}
\label{apdx:qp}
Our MPC problem stated in Equations (\ref{eq:mpc_cost}-\ref{eq:mpc_state_constraints}) is non-linear.  In order to efficiently find a solution we linearize the problem about an initial solution, and solve iteratively, in a sequential quadratic programming (SQP) fashion \cite{nocedal2006numerical}.  Let $\{\hat{x}_k, \hat{u}_k\}_{k=0,\cdots,T}$ denote an initial solution.  Let $\{\delta x_k, \delta u_k\}_{k=0,\cdots,T}$ denote deviation from the initial solution.  We approximate (\ref{eq:mpc_cost}-\ref{eq:mpc_state_constraints}) by a problem with quadratic costs and linear constraints with respect to $\{\delta x, \delta u\}$:
\begin{align}
 \{\delta x^*, \delta u^*\} &= \argmin_{\delta x, \delta u}
    \sum_{k=0}^T \|\hat{x}_{k} + \delta x_k - x^*_{k}\|_{Q_k} \nonumber\\
    &\qquad \qquad \qquad + \lambda J(\hat{x}_k + \delta x_k, \hat{u}_k + \delta u_k)
    \label{eq:mpc_linear_start}\\
 s.t. \quad \forall k&\in[0,\cdots,T]:\nonumber
 \end{align}
 \begin{align}
    \hat{x}_{k+1} + \delta x_{k+1} &= f(\hat{x}_k, \hat{u}_k) + \nabla_x f\cdot\delta x_k + \nabla_u f\cdot\delta u_k \label{eq:mpc_linear_constraints_start}\\ 
    g(\hat{u_k}) &+ \nabla_u g\cdot\delta u_k \succ 0 \\ 
    h(m,\hat{x}_k) &+\nabla_x h\cdot\delta x_k \succ 0 
    \label{eq:mpc_linear_end}
\end{align}
where $J(\hat{x}_k + \delta x_k, \hat{u}_k + \delta u_k)$ can be approximated with a second-order Taylor approximation (for now, assume no dependence on controls):
\begin{align}
    J(\hat{x} + \delta x) \approx J(\hat{x}) + \nabla_x J\cdot\delta x + \delta x^\intercal H(J) \delta x
\end{align}
and $H(\cdot)$ denotes the Hessian.  The problem is now a quadratic program (QP) with quadratic costs and linear constraints.  To solve Equations (\ref{eq:mpc_linear_start}-\ref{eq:mpc_linear_end}), we introduce the solution vector variable $X$:
\begin{equation}
    X = 
    \begin{bmatrix}
        \delta x_0^\Tr & \cdots & \delta x_T^\Tr & \delta u_0^\Tr & \cdots & \delta u_T^\Tr\\
    \end{bmatrix}^\Tr
\end{equation}
We can then write Equations (\ref{eq:mpc_linear_start}-\ref{eq:mpc_linear_end}) in the form:
\begin{align}
    \text{minimize}\quad & \frac{1}{2}X^\Tr P X + q^\Tr X\\
    \text{subject to}\quad & l\le AX \le u
\end{align}
where $P$ is a positive semi-definite weight matrix, $q$
is a vector to define the first order term in the objective function, $A$ defines inequality constraints and $l$ and $u$ provide their lower and upper limit. 

\subsection{Linesearch Algorithm for SQP solution refinement}
\label{apdx:linesearch}

The solution to the SQP problem returns an optimized control sequence $\{u_k^*\}_{k=0:T}$.  We then use a linesearch routine to find an appropriate correction coefficient $\gamma$, using Algorithm \ref{alg:linesearch}. The resulting correction coefficient is carried over into the next path-planning loop.
\vspace{5mm}

\begin{algorithm}
    \caption{Linesearch Algorithm}
    \begin{algorithmic}[1]
        \renewcommand{\algorithmicrequire}{\textbf{Input:}}
        \renewcommand{\algorithmicensure}{\textbf{Output:}}
        \REQUIRE candidate control sequence $\{u^\mathrm{c}_k\}_{k=0:T}$, QP solution $\{\delta u^*_k\}_{k=0:T}$
        \ENSURE  correction coefficient $\gamma$
        \\ \textit{Initialization}
        \STATE initialize $\gamma$ by default value or last-used value
        \STATE $[c, o]=$getCostAndObstacles($\{u^\mathrm{c}_k\}_{k=0:T}$)
        \\ \textit{Linesearch Loop}
        \FOR {$i = 0$ to $\text{max\_iteration}$}
        \FOR {$k = 0$ to $T$}
        \STATE $u^{\mathrm{c}(i)}_k=u^\mathrm{c}_k + \gamma \delta u^*_k$
        \ENDFOR
        \STATE $[c^{(i)}, o^{(i)}]=$getCostAndObstacles($\{u^{\mathrm{c}(i)}_k\}_{k=0:T}$)
        \IF {($c^{(i)} \le c$ and $o^{(i)} \le o$)}
        \STATE $\gamma=\min(2\gamma,\gamma_{max})$
        \BREAK
        \ELSE
        \STATE $\gamma=\max(\gamma/2,\gamma_{min})$
        \ENDIF
        \ENDFOR
        \RETURN $\gamma$ 
    \end{algorithmic} 
    \label{alg:linesearch}
\end{algorithm}

\end{document}